\documentclass[12pt,a4paper]{article}

\usepackage{booktabs} 

\usepackage[ruled]{algorithm2e} 

\SetAlFnt{\small}
\SetAlCapFnt{\small}
\SetAlCapNameFnt{\small}
\SetAlCapHSkip{0pt}
\IncMargin{-\parindent}

\usepackage{amsfonts}
\usepackage{amsmath}
\newcommand{\mat}[1]{\mathbf{#1}}
\renewcommand{\vec}[1]{\mathbf{#1}}

\usepackage{multirow,tabularx}
\usepackage{array}    
\usepackage{booktabs} 

\usepackage[utf8]{inputenc}

\usepackage{datatool}
\usepackage{siunitx}
\usepackage{subfigure}

\usepackage{array}
\newcolumntype{P}[1]{>{\centering\arraybackslash}p{#1}}

\usepackage{footmisc}

\usepackage{authblk}

\usepackage{graphics}
\usepackage{graphicx}
\usepackage{hyperref}

\usepackage[utf8]{inputenc}
\usepackage[english]{babel}
\usepackage[square,numbers]{natbib}
  \DTLloaddb{1nndtw}{./results/trainreport_1nnDTW_cv.csv}
  \DTLloaddb{1nnRefDTW}{./results/trainreport_1nnDTWRef_cv.csv}
  
  \DTLloaddb{RefDTWsvm}{./results/trainreport_None_svm_full-squats_angles1,translations2_sfFalse.csv} 
  \DTLloaddb{RefDTWrf}{./results/trainreport_None_rf_full-squats_angles1,translations2_sfFalse.csv} 
  
  \DTLloaddb{RefDTWrfsvm}{./results/trainreport_rf_svm_full-squats_angles1,translations2_sfFalse.csv}
  
  \DTLloaddb{RefDTWrfsvmmp}{./results/trainreport_rf_svm_movement_primitives_angles1,translations2_sfFalse.csv}

\usepackage{xstring}
\newcommand*{\cres}[3]{%
    \newcommand{\content}{#1}
    \DTLgetvalue{\val}{#2}{#3}{\dtlcolumnindex{#2}{\content}}
    \DTLgetvalue{\stdev}{#2}{#3}{\dtlcolumnindex{#2}{\content stdev}}
    \num[round-mode=places,round-precision=2]{\val} \newline \textit{(\num[round-mode=places,round-precision=2]{\stdev})} 
}%
\newcommand*{\pattern}[1]{%
    \DTLgetvalue{\val}{RefDTWsvm}{#1}{\dtlcolumnindex{RefDTWsvm}{PSP}}
    \IfSubStr{\val}{archedneck}{%
        arched neck
    }
    
    \IfSubStr{\val}{feetdistancenotsufficient}{%
        feet distance not sufficient
    }
    
    \IfSubStr{\val}{hipsdonotstart}{%
        hips do not initiate movement
    }
   
    \IfSubStr{\val}{hollowback}{%
        hollow back
    }
    
    \IfSubStr{\val}{incorrectweightdistribution}{%
        incorrect weight distribution
    }
    
    \IfSubStr{\val}{kneessideways}{%
        knees tremble sideways
    }
    
    \IfSubStr{\val}{legscompletelyextendedaftersquat}{%
        legs extended at end
    }
    
    \IfSubStr{\val}{symmetry}{%
        not symmetric
    }
    
    \IfSubStr{\val}{toodeep}{%
        too deep
    }
    
    \IfSubStr{\val}{wrongmovementdynamics}{%
        wrong dynamics
    }
}%

\newcommand*{\resline}[1]{

  \pattern{\thepidx} & 
  \cres{#1}{1nndtw}{\thepidx} & 
  \cres{#1}{1nnRefDTW}{\thepidx}  & 
  \cres{#1}{RefDTWsvm}{\thepidx}  & 
  \cres{#1}{RefDTWrf}{\thepidx}  & 
  \cres{#1}{RefDTWrfsvm}{\thepidx}  & 
  \cres{#1}{RefDTWrfsvmmp}{\thepidx}  \inclinecounter
  \\\hline
}

\newcommand*{\restableheader}[1]{%
    \begin{tabular}{c} #1 \end{tabular} & 
    1NN-DTW & 
    1NN-RefDTW & 
    RefDTW-SVM & 
    RefDTW-RF & 
    RefDTW-RF-SVM & 
    Segment-based RefDTW-RF-SVM  \\\hline 
}%
\newcounter{pidx}
\newcommand{\inclinecounter}{%
        \stepcounter{pidx}}

\begin{document}
\begin{filecontents*}{./results/trainreport_1nnDTW_cv.csv}
PSP,f1,f1stdev,accuracy,accuracystdev,roc,rocstdev
archedneck,0.7002997003,0.110212948717,0.695687645688,0.0888784626114,0.693333333333,0.0842386201496
feetdistancenotsufficient,0.632535885167,0.081157477431,0.55981092437,0.113285243345,0.539642857143,0.128240338935
hipsdonotstart,0.15873015873,0.223578626815,0.529642857143,0.148347091008,0.426818181818,0.149740822923
hollowback,0.584731058415,0.0988100738195,0.606904761905,0.057736990662,0.608531746032,0.0654358942861
incorrectweightdistribution,0.72825311943,0.1457126306,0.618974358974,0.204740659549,0.57696969697,0.273086295674
kneessideways,0.313333333333,0.0905811143208,0.46,0.0719567771497,0.438333333333,0.0659072391497
legscompletelyextendedaftersquat,0.541415839063,0.177599280749,0.6175,0.147903749213,0.62380952381,0.143944823175
symmetry,0.383174603175,0.199973795209,0.558058608059,0.182382640783,0.542222222222,0.196185852927
toodeep,0.509431153642,0.134187725901,0.472058823529,0.139543248991,0.468787878788,0.142282617515
wrongmovementdynamics,0.769071225071,0.0810197777179,0.653792569659,0.12581330655,0.546806526807,0.174946179569
\end{filecontents*}

\begin{filecontents*}{./results/trainreport_1nnDTWRef_cv.csv}
PSP,f1,f1stdev,accuracy,accuracystdev,roc,rocstdev
archedneck,0.621616161616,0.187945681585,0.662121212121,0.130740141943,0.66380952381,0.12794733894
feetdistancenotsufficient,0.642188805347,0.106300861644,0.555091036415,0.0992833118286,0.525396825397,0.0899525364053
hipsdonotstart,0.226666666667,0.137275068546,0.586666666667,0.135806985884,0.472727272727,0.105949075558
hollowback,0.535164835165,0.101408984572,0.59369047619,0.0808346482816,0.589880952381,0.0869618207988
incorrectweightdistribution,0.695993179881,0.0759410413327,0.569487179487,0.107589694703,0.506060606061,0.185933230509
kneessideways,0.461587301587,0.212462318625,0.566666666667,0.156702123647,0.556904761905,0.160134580588
legscompletelyextendedaftersquat,0.609523809524,0.20102345843,0.615098039216,0.181259212072,0.613888888889,0.178184545627
symmetry,0.327777777778,0.232006811309,0.554395604396,0.19295686696,0.496666666667,0.196059955557
toodeep,0.490952380952,0.170276552433,0.483169934641,0.157287347448,0.493116883117,0.161602282438
wrongmovementdynamics,0.754971786834,0.065117499942,0.652321981424,0.103238588964,0.580291375291,0.149647401375
\end{filecontents*}

\begin{filecontents*}{./results/trainreport_None_svm_full-squats_angles1,translations2_sfFalse.csv}
PSP,f1,f1stdev,accuracy,accuracystdev,roc,rocstdev
archedneck,0.668787878788,0.0753297007827,0.632634032634,0.09300202708,0.719365079365,0.121332153857
feetdistancenotsufficient,0.981818181818,0.0363636363636,0.976470588235,0.0470588235294,0.994285714286,0.0114285714286
hipsdonotstart,0.411587301587,0.311927823373,0.715595238095,0.136031583968,0.710363636364,0.115178051169
hollowback,0.645987345987,0.236527358776,0.7075,0.194550764583,0.702976190476,0.21580480073
incorrectweightdistribution,0.904775707384,0.0728476525275,0.857509157509,0.111384226051,0.947727272727,0.047145321709
kneessideways,0.374141414141,0.305606290136,0.51,0.219494368442,0.439761904762,0.309378720574
legscompletelyextendedaftersquat,0.572128851541,0.211123478957,0.605,0.107975142027,0.641121031746,0.132160768186
symmetry,0.35,0.226077666104,0.68021978022,0.119531841369,0.615555555556,0.126889559143
toodeep,0.816031746032,0.144864555641,0.780964052288,0.168216633003,0.879480519481,0.126283994069
wrongmovementdynamics,0.913428571429,0.0810712852247,0.88363003096,0.105420371808,0.919137529138,0.0872927352396
\end{filecontents*}

\begin{filecontents*}{./results/trainreport_None_rf_full-squats_angles1,translations2_sfFalse.csv}
PSP,f1,f1stdev,accuracy,accuracystdev,roc,rocstdev
archedneck,0.754062604063,0.062012682466,0.729020979021,0.0696673386195,0.721904761905,0.0701860178205
feetdistancenotsufficient,0.95037593985,0.0522218206141,0.939705882353,0.0644631141009,0.932857142857,0.0705083581672
hipsdonotstart,0.697619047619,0.0708708384102,0.836904761905,0.036685214851,0.78,0.0430116263352
hollowback,0.938461538462,0.0897069522284,0.948214285714,0.0732795876024,0.947817460317,0.0770178702027
incorrectweightdistribution,0.963333333333,0.0452155332208,0.942564102564,0.0706427706983,0.906666666667,0.114794502385
kneessideways,0.0444444444444,0.0888888888889,0.486666666667,0.0678232998313,0.417619047619,0.0481305857847
legscompletelyextendedaftersquat,0.932679738562,0.0389621433417,0.926274509804,0.0433435353087,0.924603174603,0.0450488592847
symmetry,0.0,0.0,0.697985347985,0.0614069543744,0.477777777778,0.0444444444444
toodeep,0.905614035088,0.0758609149967,0.873202614379,0.11173221253,0.856623376623,0.139450164778
wrongmovementdynamics,0.953634018156,0.027424581623,0.93363003096,0.0387356643944,0.902307692308,0.0545762469616
\end{filecontents*}

\begin{filecontents*}{./results/trainreport_rf_svm_full-squats_angles1,translations2_sfFalse.csv}
PSP,f1,f1stdev,accuracy,accuracystdev,roc,rocstdev
archedneck,0.720670995671,0.0784616360494,0.696736596737,0.106316196042,0.753174603175,0.109982934612
feetdistancenotsufficient,0.960765550239,0.0349511624263,0.951470588235,0.0444472978227,0.985079365079,0.0186086995425
hipsdonotstart,0.588095238095,0.161413821492,0.784404761905,0.0635411020941,0.862636363636,0.0463443810283
hollowback,0.89619047619,0.109519668659,0.908095238095,0.0949949276603,0.96626984127,0.0493405747377
incorrectweightdistribution,0.961111111111,0.0484322104838,0.942564102564,0.0706427706983,1.0,0.0
kneessideways,0.362121212121,0.264019227393,0.553333333333,0.187794213613,0.637619047619,0.264991122086
legscompletelyextendedaftersquat,0.770115995116,0.0660746985258,0.762450980392,0.0727419268903,0.877083333333,0.0571831455196
symmetry,0.313333333333,0.269649978882,0.710989010989,0.149727487074,0.591296296296,0.0956344778873
toodeep,0.870882740448,0.0754230007948,0.849673202614,0.0857181506445,0.917922077922,0.0714092063478
wrongmovementdynamics,0.934586399108,0.0195023686369,0.909365325077,0.0269549941689,0.961608391608,0.0262950701427
\end{filecontents*}

\begin{filecontents*}{./results/trainreport_rf_svm_movement_primitives_angles1,translations2_sfFalse.csv}
PSP,f1,f1stdev,accuracy,accuracystdev,roc,rocstdev
archedneck,0.724300699301,0.0736178918709,0.712121212121,0.0987255854137,0.783968253968,0.151973993611
feetdistancenotsufficient,0.98,0.04,0.975,0.05,1.0,0.0
hipsdonotstart,0.667619047619,0.0740257968151,0.808333333333,0.0587762581837,0.898454545455,0.0259108452357
hollowback,0.914139194139,0.105008444759,0.922380952381,0.0915673549845,0.967261904762,0.0556475994882
incorrectweightdistribution,1.0,0.0,1.0,0.0,1.0,0.0
kneessideways,0.509668109668,0.217510438509,0.573333333333,0.18636880998,0.654761904762,0.225188255837
legscompletelyextendedaftersquat,0.915072900955,0.0221364364075,0.904761904762,0.031153194731,0.978571428571,0.0285714285714
symmetry,0.426031746032,0.132063492063,0.668181818182,0.0982397883482,0.653240740741,0.182118119917
toodeep,0.838658988071,0.102628328223,0.81364379085,0.084089387161,0.923116883117,0.0620034330031
wrongmovementdynamics,0.910805860806,0.0405734498535,0.877283281734,0.048358263491,0.931398601399,0.0332691771794
\end{filecontents*}

\title{Automatic Error Analysis of Human Motor Performance for Interactive Coaching in Virtual Reality}  
\date{}
\author[1,2]{Felix Hülsmann}

\author[2]{Stefan Kopp}
\author[1]{Mario Botsch}
\affil[1]{Computer Graphics Group}
\affil[2]{Social Cognitive Systems Group}

\affil[  ]{Bielefeld University, Germany}

\maketitle
\begin{abstract}
In the context of fitness coaching or for rehabilitation purposes, the motor actions of a human participant must be observed and analyzed for errors in order to provide effective feedback. This task is normally carried out by human coaches, and it needs to be solved automatically in technical applications that are to provide automatic coaching (e.g. training environments in VR). However, most coaching systems only provide coarse information on movement quality, such as a scalar value per body part that describes the overall deviation from the correct movement. Further, they are often limited to static body postures or rather simple movements of single body parts.
While there are many approaches to distinguish between different types of movements (e.g., between walking and jumping), the detection of more subtle errors in a motor performance is less investigated. We propose a novel approach to classify errors in sports or rehabilitation exercises such that feedback can be delivered in a rapid and detailed manner: Homogeneous sub-sequences of exercises are first temporally aligned via Dynamic Time Warping. Next, we extract a feature vector from the aligned sequences, which serves as a basis for feature selection using Random Forests. The selected features are used as input for Support Vector Machines, which finally classify the movement errors. We compare our algorithm to a well established state-of-the-art approach in time series classification, 1-Nearest Neighbor combined with Dynamic Time Warping, and show our algorithm's superiority regarding classification quality as well as computational cost.

\end{abstract}

\section{Introduction}
Coaching environments for motor learning have become a more and more popular research topic in the field of Virtual Reality (VR) \cite{kyan2015approach,de2015multimodal,sigrist2015sonification,chan2011virtual}. They are promising in areas such as rehabilitation or fitness training. Obviously, high-quality feedback on the coachee's performance is crucial for the success of such systems. Therefore, an intelligent coaching system does not only have to detect which task --- in the following called \textit{motor action} --- is executed. It also has to detect the specific errors the coachee performs during an exercise and has to address them using appropriate feedback. While lots of approaches exist for the classification of motor actions~\cite{endres2016bayesian,giggins2014use,chen2016online,ahmadi2015toward,um2016exercise,hosseini2015efficient}, 
fewer consider the analysis of the performance quality. If they do, authors often focus on reporting simple scores, which summarize the performance quality in terms of a deviation from a desired performance~\cite{chan2011virtual,sun2016assessment,kyan2015approach}. Others provide scoring functions which describe overall improvement or decline in quality for a specific exercise~\cite{houmanfar2014movement}. However, many types of complex sports movements can be executed correctly yet with different individual styles~\cite{hossner2015functional}. Moreover, some parts of the body are often completely irrelevant for the successful execution of the movement. For instance, the orientation of the hands is negligible when analyzing the quality of a body weight squat. Consequently, feedback that only relies on an overall deviation from a prerecorded desired performance, including task-irrelevant deviations, is non-optimal when aiming at improving the coachee's performance~\cite{sigrist2013augmented,liu2007evidence}.

For many types of motor actions, a set of typical errors can be found~\cite{bartlett2007introduction}. 
Often, there is only a very subtle distinction between a correct movement and the occurrence of a certain error. 
For many known errors, coaches have established feedback strategies to support a coachee in improving her performance. This could be, for instance,  verbal descriptions of the error together with best practices on how to eliminate it. Intelligent coaching environments in VR need to be able to detect such error patterns automatically and to provide elaborate feedback, e.g. taken from real-world coaching experience. Such feedback must be provided online or rapidly, i.e., either directly after a coachee has finished the movement or --- even better --- already during the motor action being performed. Some approaches try to achieve this using manually designed rules that can be evaluated online~\cite{de2015multimodal,rector2013eyes,hulsmann2016multi}. However, this requires enormous manual effort and bears the risk of gaps or under-fitting of the designed rules.

 \begin{figure} 
	\includegraphics[width=1.0\columnwidth]{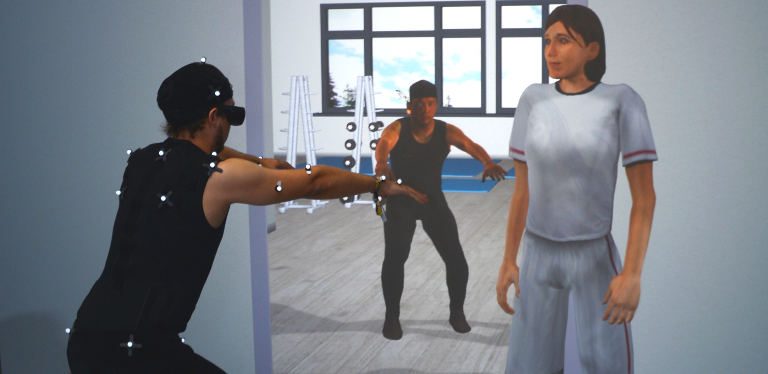}    
 \caption{In our real-time VR coaching environment, a coachee performs exercises while being observed by a virtual coach. The coach
 needs to extract information on performed errors to provide effective feedback.}
  \label{fig:teaser}
 \end{figure}

In this paper, we present an approach to automatic error analysis of human motor performance in an immersive VR coaching environment for sports and rehabilitation exercises (see Figure~\ref{fig:teaser}). We focus on the squat movement as a test case for our approach. The squat is a full-body motor action that is frequently used in the context of rehabilitation~\cite{bailey2011single,escamilla2001knee} as well as for sports training~\cite{escamilla2001knee}. When executed by novice coachees, various different error patterns can be observed in a squat. 
We consider the detection of such error patterns as a time series classification problem. In the field of time series classification, 1-Nearest-Neighbor combined with Dynamic Time Warping (1NN-DTW) proved to be state of the art and difficult to beat by other classifiers~\cite{xi2006fast,bagnall2014experimental}. 
We aim to extend the current state of the art in the classification of typical error patterns in motor performance. Our contribution is as follows:
\begin{itemize}
 \item We propose a novel approach towards the classification of error patterns in motor performances which uses a reference-based Dynamic Time Warping of movement segments as a basis for a feature selection using Random Forest. The selected features are in a final step classified by a Support Vector Machine (SVM).
 \item We show that this classifier outperforms the 1NN-DTW approach, in both classification performance as well as time needed for classification.
 \item We show the effectiveness of the approach on an exemplary data set and demonstrate the impact of all components on classification performance as well as on time needed for classification. 
\end{itemize}

In the next section, we discuss related work towards motor performance analysis and time series classification. Then, we describe how we obtain our data set, which consists of a list of typical error patterns, together with annotated movement data. In Section~\ref{sec:algorithm}, we first evaluate the performance of 1NN-DTW on our data set. Next, we provide a step-by-step evaluation of the components of our approach. 
In Section~\ref{sec:discussion}, we discuss the results and conclude the paper. The video in the online material demonstrates how we use the proposed analysis to generate verbal feedback inside our ``Intelligent Coaching Space''\footnote{\url{http://graphics.uni-bielefeld.de/research/icspace/}}, an immersive coaching environment for sports and rehabilitation exercises (see Figure~\ref{fig:teaser})~\cite{waltemate2015realizing}.

\section{Related Work}
\label{sec:rw}
Two main approaches have been applied to assess the quality of human motor performances. The first approach (Section~\ref{sec:rwA}) is to engineer a highly specialized method, e.g., for the evaluation of feedback strategies for a very specific type of motor action. In this approach, a common choice is to assess quality by determining the overall distance of the performed motion to the desired motion. Often, a model for these specific performance patterns is manually designed drawing from expert knowledge. The second direction (Section~\ref{sec:rwB}) consists in using more general, data-based approaches, such as well established techniques from time series classification. In the following, we will present and discuss work stemming from both directions.

\subsection{Specific, Manually Designed Approaches}
\label{sec:rwA}
\citeauthor{houmanfar2014movement} use a manually designed scoring function to represent patients' performance changes in a rehabilitation setting~\cite{houmanfar2014movement}. Even though this approach provides compelling results in the field of application, no detailed information on occurred error patterns is gained, which would be necessary for the application of complex coaching strategies. 

Other approaches make use of rule-based systems to detect the occurrence of certain error patterns. In the context of yoga training, \citeauthor{rector2013eyes} define optimal yoga poses~\cite{rector2013eyes}. De Kok et al. went one step further by manually defining error patterns~\cite{de2015multimodal} focussing on the whole trajectory. Rules are implemented, first to split the motion into sequential movement segments, and then to describe the error patterns. A state machine performs the classification. 

One major advantage of the approaches by Rector et al. or de Kok et al. is their real-time capability: Specific feedback strategies linked to typical error patterns can be applied immediately. Further, the results are deterministic: If the rules are correct and exhaustive and the motion capture system works properly, an incorrect classification is unlikely to occur. This directly leads to the major disadvantage: As the rules have to be designed manually, they are prone to errors during the design phase, which might be difficult to be tracked down later on. A single error during the design of only one pattern might have a devastating effect on the resulting system in terms of effectiveness and even safety of the training. Moreover, it is mostly not trivial --- even when interviewing sports coaches --- to obtain exact information about which features are significant or where to draw the border between a correct or an incorrect movement. Finally, the design of rules requires enormous manual effort: For each motor action and for each type of error, a detailed investigation on how to describe the motor action and the error has to be performed. For complex error patterns, this quickly becomes infeasible. Thus, it is desirable to focus on approaches that automatically learn most of their information from data.

\subsection{Data-based Approaches}
\label{sec:rwB}
\citeauthor{taylor2010classifying} focus on classifying error patterns in rehabilitation exercises using a combination of rule-based segmentation and AdaBoost on a set of manually defined features~\cite{taylor2010classifying}.  In a within-subject cross validation, the authors obtain highly convincing results. However, classification performance decreases significantly when generalizing to new subjects. Furthermore, the design of feature sets requires additional manual work.

\citeauthor{kianifar2016classification} present an approach towards distinguishing between good, moderate, and bad performances of squat movements~\cite{kianifar2016classification}. They use a feature vector based on manually designed features, such as skewness and range, whose dimensionality is reduced using Sparse Principal Component Analysis (SPCA). Finally, Decision Trees are used for classification. The classification accuracy to distinguish between good, moderate, and bad squats in a leave-one-subject-out cross validation is \SI{73}{\%}. For the distinction between only two classes (good and bad), a higher accuracy of \SI{98.6}{\%} was achieved. The presented approach is only able to distinguish between three coarse classes of quality and cannot spot single error patterns. In addition, manual effort is needed for feature preparation. Furthermore, SPCA is an unsupervised algorithm, which searches for a set of sparse principal components which cover as much as possible of the variance inside the data \cite{zou2006sparse}. This is problematic when most of the variance is due to individual differences rather than performance errors, which holds for sports movements that can differ considerably between subjects.

\citeauthor{o2015evaluating} use a neural network classifier to differentiate between correct and incorrect performances of squats and to classify error patterns. A leave-one-out cross validation resulted in an accuracy of \SI{80}{\%} to distinguish between correct and incorrect, but only in an accuracy of \SI{57}{\%} for the classification of error patterns. Similar experiments were conducted by \citeauthor{giggins2013evaluating}~\cite{giggins2013evaluating,giggins2014rehabilitation}.

\citeauthor{yurtman2014automated} proposed an extension of Dynamic Time Warping (DTW) that is able to detect multiple occurrences of multiple exercise types in trajectories as well as to classify error patterns~\cite{yurtman2014automated}. Classification is performed by comparing the just performed motion to pre-recorded templates and then selecting the best matching one. This leads to a very high accuracy of \SI{93}{\%} for exercise classification and \SI{89}{\%} for the classification of errors in motor performances (inter-subject performance was not tested). 
However, combinations of multiple error patterns cannot be considered as long as they are not included as individually pre-recorded templates.

Overall, the data-based approaches employed in the context of sports and rehabilitation applications have three weaknesses: First, it is often not analyzed how well the trained classifiers generalize to new subjects. Many approaches require the system be re-trained for each user. This leads to problems as subjects are often physically not able to provide all the required training data. For instance, in the context of sports performances, some users are not able to perform the desired motor action correctly or, on purpose, with a certain type of error. Second, the motor actions and error patterns are often rather simple. Some of the presented systems only distinguish between, e.g., ``good'' or ``bad'' for a motor action that only involves a very small number of joints. Especially algorithms using variance-based dimensionality reduction or pure comparisons with prototypes will perform worse on more subtle errors or more complex movements: Most of the variance and also the similarity to prototypes would be covered by inter-subject variations instead of the movement patterns underlying the errors. Finally, for most algorithms, no information on the applicability in interactive or real-time systems is given. Especially algorithms which require expensive calculations for each classification do not meet the requirements of VR coaching systems as, e.g., described in \cite{waltemate2015realizing}.

Another group of data-based approaches has been developed in the field of Computer Graphics to capture and synthesize human motion with particular styles. Analysis of  observed movements is then often possible through ``analysis by synthesis''.
Giese et al.~introduced Spatio-Temporal Morphable Models for analysis and synthesis of morphs between gait styles~\cite{giese2000morphable}. First, recordings of prototypical performances are brought into spatio-temporal correspondence. Then, new trajectories can be described as spatio-temporal blends between prototypes. The underlying assumption is that a clearly defined prototype can be obtained for each desired style. In our case, these styles would be the possible error patterns in a motor performance. However, in the context of motor learning, movements often contain a combination of different error patterns and prerecorded single prototypical errors do not work equally well for different subjects. 

A related approach has been proposed by \citeauthor{min2012motion}~\cite{min2012motion}: Their model, called Motion Graphs++, describes human movements by (a) discrete structural variations that define the motor action together with (b) continuous variations that capture the movement style.
Style variations are represented using Principal Component Analysis (PCA) together with a Mixture of Gaussians. MotionGraphs++ are powerful as they do not need an isolated demonstration of each prototype. However, if a targeted variation in style is not covered by the PC dimensions, the model cannot detect this style pattern. In the case of typical error patterns in motor performances, the differences between users who perform the same error may be  relatively big, whereas the difference between error patterns within a user can be very subtle. Thus, MotionGraphs++ would rather encode the inter-individual differences than the characteristics of the error patterns.

Finally, the classification of errors in motor performances is a special case of time series classification, for which several machine learning algorithms have been proposed. Ground-breaking work was performed by \citeauthor{wilson1999parametric}, who used hidden Markov models (HMM) for the recognition of gestures~\cite{wilson1999parametric}. Other methods are based on decision trees~\cite{rodriguez2004interval}, SVMs~\cite{wu2004distance}, or Multi-Layer Perceptrons (MLP)~\cite{nanopoulos2001feature}. Dynamic Time Warping (DTW) is usually used to temporally align two recorded trajectories. As a pseudo-metric combined with a subsequent classification, DTW has a highly positive impact on motion classification~\cite{adistambha2008motion,petitjean2014dynamic,xi2006fast}. Xi et al.~provide an extensive review comparing a large set of available classification methods, such as HMMs, MLPs, and decision trees on time series data~\cite{xi2006fast}. They show that no tested classifier is able to beat a combination of DTW and 1-Nearest-Neighbor (1NN-DTW), which basically compares the query trajectory to each available training trajectory using DTW as distance measure. Then the most similar training trajectory is used to predict the label of the query trajectory. The superiority of this approach in comparison with nine classifiers, including Random Forests, SVM, Bayes Networks, et cetera, is supported by work from \citeauthor{bagnall2014experimental} \cite{bagnall2014experimental}. Likewise, \citeauthor{yurtman2014automated} achieved good classification results using a method similar to 1NN-DTW, which, however, was limited to simple movement patterns and was not evaluated with respect to generalization to movements of other persons~\cite{yurtman2014automated}. 
 
To sum up, the approaches discussed in this section suffer from a number of limitations that prevent their use for real-time coaching of human motor performances. We aim to go beyond this by developing a classification approach that can classify subtle errors in a complex motor action with high accuracy, works on a small or unbalanced dataset, achieves good generalization over different users, and provides its results very quickly and already after relevant parts of the performance have been observed. We will base our approach on knowledge from Sports Science about which errors are particularly relevant, and we present an approach that determines discriminatory features of these errors and then realizes classifiers with the desired properties. We will take 1NN-DTW as a baseline in evaluating them. 

\section{Domain and Dataset}
\label{sec:dataset}
\begin{figure}[] 
 \centering
	\includegraphics[width=0.25\columnwidth]{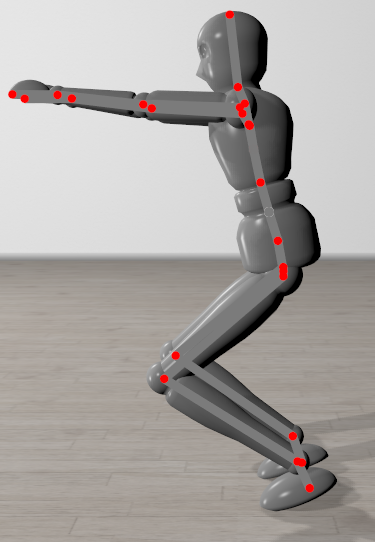} 
	 \caption{Squat performed with error pattern ``incorrect weight distribution''.}
	 \label{fig:squat}	
 \end{figure}

Sports coaches and sports scientists have developed coaching strategies to address specific error patterns during a coaching session. Before developing a VR coaching system, and to enable it to detect those errors automatically, it is important to identify relevant error patterns along with corresponding feedback strategies for each motor action of interest. To this end, we analyzed 21 video recordings of real-world squat coaching sessions. A part of these data comes from the corpus described in~\cite{de2014dialogue}; additional other videos were recorded in our lab. We used the videos together with information from Sports Scientists as well as literature (e.g.~\cite{clark2008nasm}) to compile a list of 21 relevant error patterns. For instance, one error pattern is an incorrect weight distribution (depicted in Figure~\ref{fig:squat}), which happens if the coachee shifts major parts of the body weight too much to the front. 

\begin{figure}[] 
	 \centerline{ 
		\subfigure[41 Markers placed on the subject's body.]{\label{fig:markersetup}
		\includegraphics[width=0.35\columnwidth]{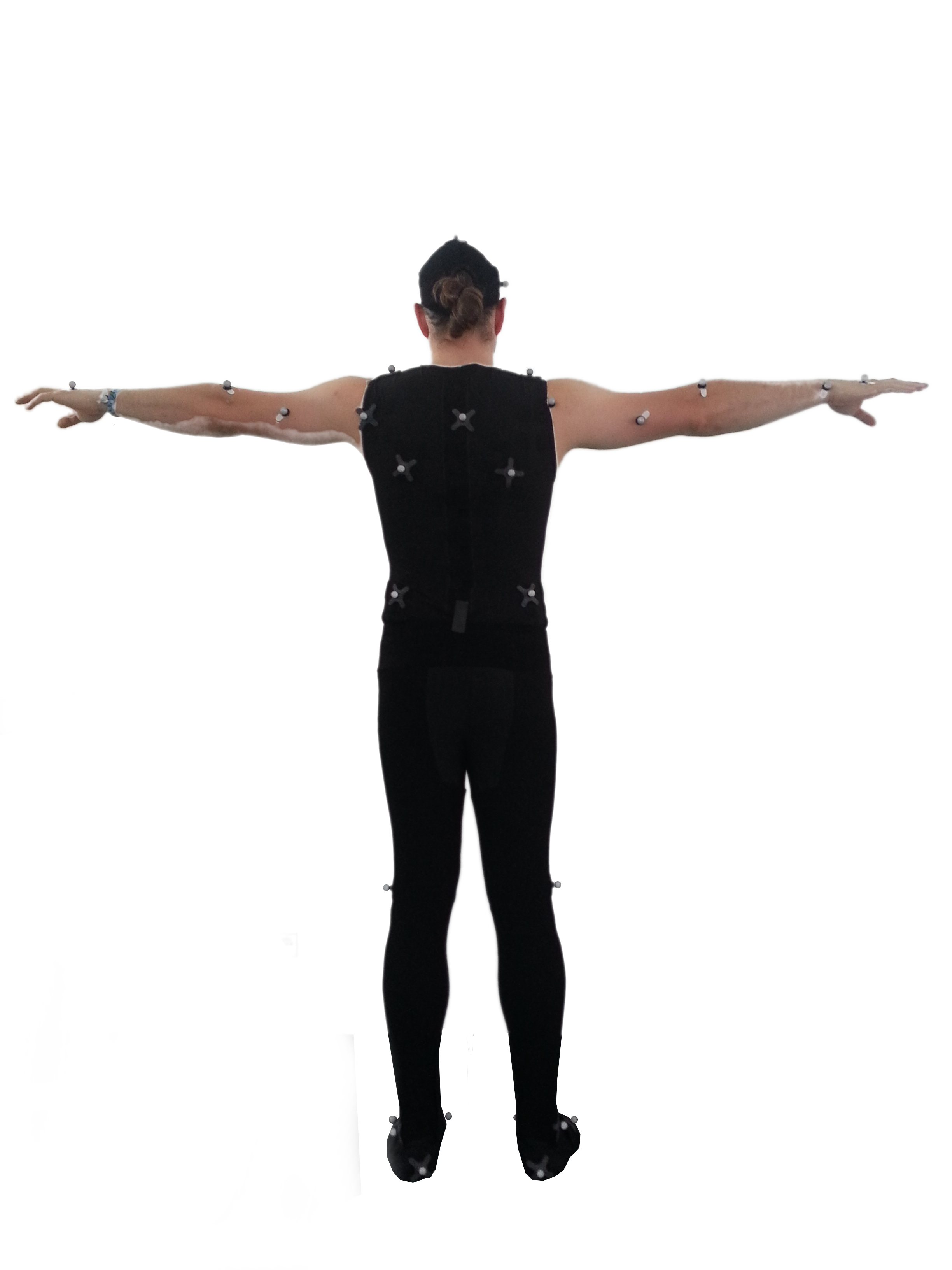}}\hfill
		\subfigure[Skeleton reconstructed from the 41 markers. We use the hips as root joint.]{\label{fig:joints}
		\includegraphics[width=0.4\columnwidth]{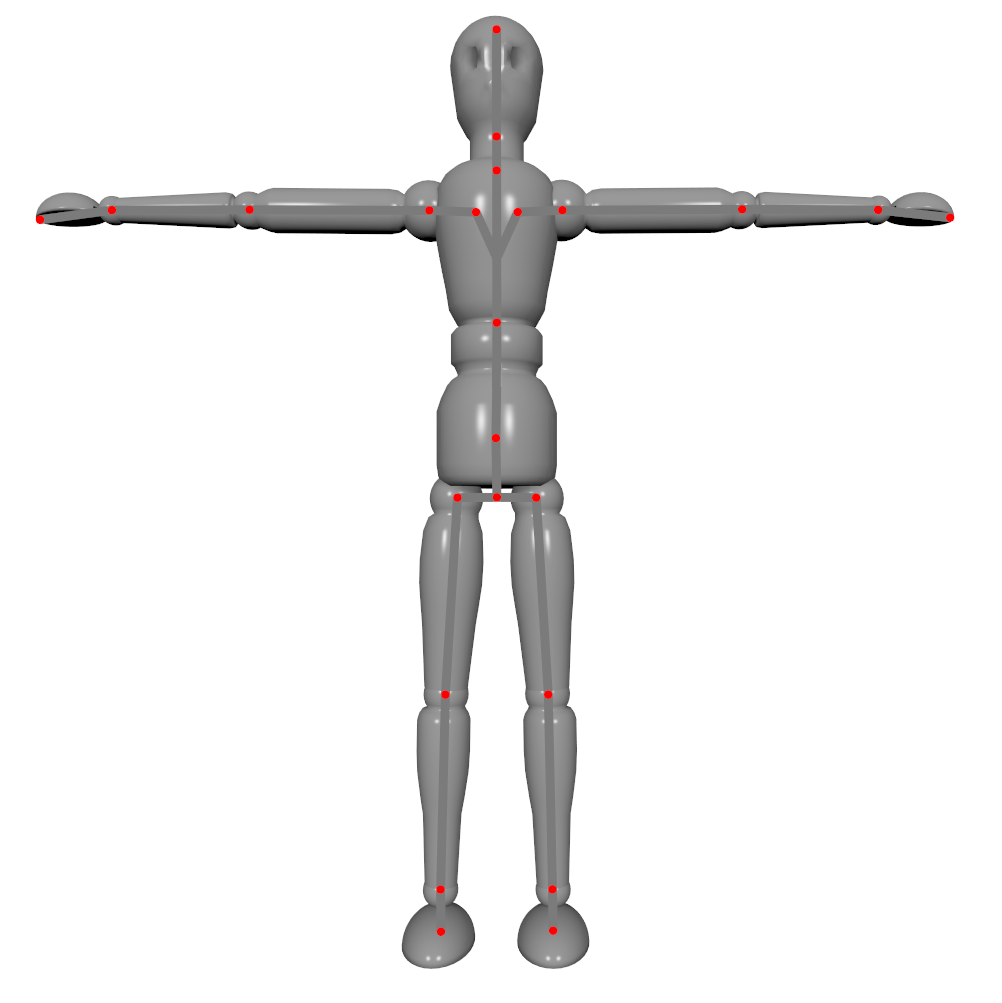}}} 
	 \caption{Marker setup and reconstructed skeleton representation.}
 \end{figure}

Motion data was recorded using an OptiTrack motion capture system, which consists of ten Prime 13W cameras. Passive markers were mostly attached to a customized motion capture suit; markers at the arms and the hands were directly attached to the subjects' skin (see Figure~\ref{fig:markersetup}). The motion capture system outputs kinematic features for 19 joints (see Figure~\ref{fig:joints}) per frame at \SI{120}{Hz}. In our representation, each frame consists of $k$ joint rotations as well as $k$ joint positions (with $k=19$). Joint rotations are represented as quaternions $\vec{q}_1, \dots, \vec{q}_k$. Each quaternion denotes the rotation of a joint with respect to its parent. The root rotation $\vec{q}_1$  describes to rotation of the root with respect to its rotation at the beginning of the movement. As root joint, we use the hips. The joint positions are represented by vectors $\vec{t}_1, \dots, \vec{t}_k \in \mathbb{R}^3$. Each denotes the y- component of the translation (height) of the joint as well as the translation relative to the x- and z- position of the root joint at the beginning of the movement, after removing the subjects orientation at the beginning of the movement. Further we additionally use joint angles as Euler angles, calculated from the quaternion representation, which correspond to flection/extension, abduction/adduction and twist of the corresponding joint.

We asked 49 subjects to perform squats inside the capture volume. Up to two squats per participant were annotated by an expert for the presence of any of the error patterns. The expert had to add confidence and intensity ratings for each decision. These ratings were combined into a score in the interval $[0,1]$ by averaging. Only ratings with a score above $0.5$ were used for the experiment. Trajectories which contained severe errors caused by the motion capture system (e.g. due to missing markers), were excluded. The final training data set consisted of $N=95$ squat movements coming from 49 subjects. We selected the error patterns that appeared with a sufficient frequency (at least 15 positive and negative examples) for training. The ten resulting patterns and their frequency in the training data are listed in Table~\ref{tab:mpeps}.

\begin{table}%
\caption{Possible error patterns in the execution of a complex squat motor action, selected based on their relevance as well as frequency and saliency in our motion data (rated by sports experts).}
\label{tab:mpeps}
\begin{minipage}{\columnwidth}
\begin{center}
\begin{tabular}{lP{3.1cm}P{3.1cm}}
  \toprule
Performance Error Pattern        		& \#Erroneous Executions 	& \#Correct Executions\\\hline
arched neck     						& 33 						& 29 \\\hline
feet distance not sufficient     		& 45						& 33 \\\hline
hips do not initiate movement			& 23		 				& 51 \\\hline
hollow back     						& 34						& 42 \\\hline
incorrect weight distribution 			& 51	 					& 16 \\\hline
knees tremble sideways     				& 23	 					& 33 \\\hline
legs extended at end                    & 42	 					& 38 \\\hline
not symmetric 							& 17						& 46 \\\hline
too deep     							& 51	 					& 34 \\\hline
wrong dynamics               			& 61	 					& 27 \\\hline
\end{tabular}
\end{center}
\end{minipage}
\end{table}%

\section{Classification Algorithm}
\label{sec:algorithm}
The combination of Dynamic Time Warping and 1-Nearest-Neighbor (1NN-DTW) is one of the most successful classifiers for time series classification~\cite{xi2006fast,bagnall2014experimental}. Thus it will serve as our baseline. In the following, we first report how we evaluate classifier performance. Then we describe the 1NN-DTW baseline approach and carve out its drawbacks for motor performance analysis in interactive coaching sessions. Then, we develop classifiers to eliminate or mitigate its weaknesses step by step. Finally, we verify that our approach is suitable for error analysis of human motor performances in the context of interactive VR coaching sessions. 

\subsection{Evaluation Procedure}
Motor actions in sports or rehabilitation training often exhibit large inter-subject variation~\cite{hossner2015functional}. Consequently, it is important to ensure that classifiers are tested on data from persons whose performances are not included in the training data. 
This hypothesis is experimentally supported by \citeauthor{taylor2010classifying}, who measure a huge difference in classifier scores when testing on samples from participants included in the training set, as compared to samples from participants who were not included in the training data~\cite{taylor2010classifying}. We made sure that for the results described in the following, no data from subjects who provided a recording to the training set is contained in the test set. We applied 5-fold cross validation under this constraint for each error pattern. In each fold, we aimed at achieving a similar proportion of positive and negative labels as in the overall data set. For our experiments, the variables of interest are the quality of the classification and the time needed for the classification of a single query trajectory.

To investigate the quality of a classification, different types of scores can be used. We report the accuracy of the described classifier, defined as the number of correctly classified samples weighted by the overall number of samples:
\begin{equation*}
 acc = \frac{\#\!T\!P + \#\!T\!N}{\#\!P + \#\!N}.
\end{equation*}
$\#\!T\!P$ is the number of true positives and $\#\!T\!N$ the number of true negatives. $\#\!P$ is the overall number of positive examples and $\#\!N$ the overall number of negative examples in the training data. Additionally, at the end of Section~\ref{sec:algorithm}, we provide plots for F1 scores, which is the harmonic mean of precision and recall of the classifier: 
\begin{equation*}
 F1 = \frac{2\#\!T\!P}{2\#\!T\!P + \#\!F\!P + \#\!F\!N}.
\end{equation*}
Here, $\#\!F\!P$ is the number of false positives, and $\#\!F\!N$ the number of false negatives.  All measured scores and standard deviations for the cross validation folds can be found in the supplementary online material.

In addition to the quality of classification, we report information on the time each algorithm needs to  classify a new query trajectory. As DTW is an essential part for each of the proposed algorithms, we report the time that is approximately needed for a DTW without any parallelization. Furthermore, to be able to compare the algorithms that only have to perform one DTW per query, we report the average time per query needed for the classification of a single error pattern. All experiments were conducted on a machine with Intel Xeon CPU E5-1620 \SI{3.6}{Ghz}.

\subsection{Baseline: 1NN-DTW}
\label{sec:baseline}
As described above, we take as baseline one of the most successful classification algorithms for time series: 1-Nearest-Neighbor as classification algorithm together with Dynamic Time Warping as distance measure (1NN-DTW). For an input query, 1NN searches for the data point that is most similar to the input. Then it returns the classification label of this nearest neighbor in the training set. The underlying assumption is that data points that lie nearby belong to the same class. To determine which points lie nearby, a frame-wise comparison is problematic in time series such as motion trajectories. If the trajectories would be compared simply frame-to-frame, results would be highly distorted: Even if the movement is performed completely in the same way in space, but with a slight temporal offset, this measure would report a very high distance, whereas if a movement is performed with similar timing but different postures (e.g. a slightly weaker movement of some joints), the distance would be very low. Dynamic Time Warping (DTW) is typically used to solve this problem as it establishes a frame-to-frame correspondence between two trajectories by warping in time and then allows to determine the distance between them. 

We implemented 1NN-DTW as follows. Given two trajectories $T_1$ and $T_2$, consisting of $n$ and $m$ frames, respectively, we use DTW to calculate the optimal match between them~\cite{sakoe1978dynamic}. First, a $n \times m$ local cost matrix $\mat{M}$ is constructed. Each element $\mat{M}(i,j)$ of this matrix corresponds to the distance between the postures $T_1(i)$ and $T_2(j)$. This distance is defined as the sum of the quaternion distances of the corresponding joints. As quaternion distance, we use the inner product as evaluated by \citeauthor{huynh2009metrics}~\cite{huynh2009metrics}. Thus, each element in the matrix $\mat{M}$ is calculated as follows:
\begin{equation*}
 \mat{M}(i,j) = \sum_{d=1}^{k} (1- |\vec{q}_{i,d} \vec{q}_{j,d}|).
\end{equation*}
To establish a frame-to-frame correspondence, an optimal path through $\mat{M}$ from $\mat{M}(1,1)$ to $\mat{M}(n,m)$ is determined based on dynamic programming. The distance between the two trajectories $T_1$, $T_2$ can now be defined as mean value of the $\mat{M}(i,j)$ on the warping path. Comparison of classification results using different features, such as joint angles or joint positions, yielded no significant improvements in the 1NN step. Results of these comparisons can be found in the supplementary online material.

\begin{figure}[] 
 \centering
 \centerline{  
 		\includegraphics[width=1.0\columnwidth]{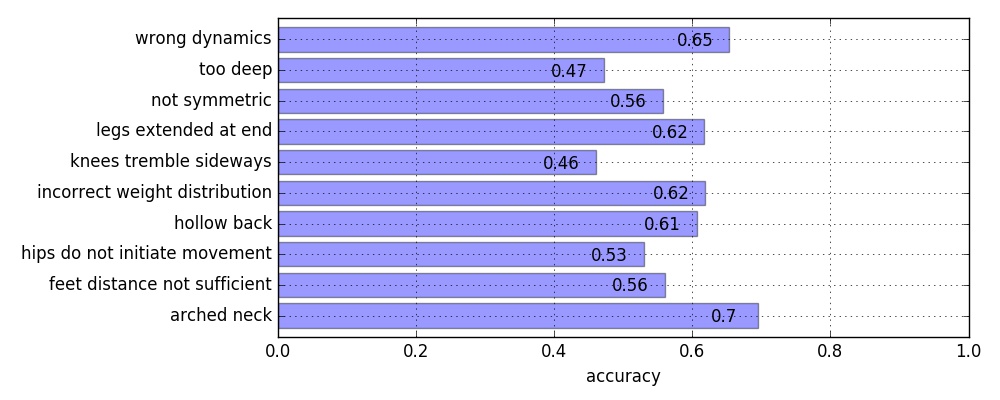}
 }  
 \caption{Accuracy results for classifier 1NN-DTW.}
 \label{fig:res:1nndtw}
 \end{figure}

We applied the above procedure to the relevant error patterns: For each query trajectory $T_q$ we compute DTW to each training trajectory $T_1, \dots, T_N$. Next, the trajectory with the smallest DTW distance to $T_q$ which is annotated with respect to the error pattern, is selected. The label of this trajectory is then returned for $T_q$. As shown in Figure~\ref{fig:res:1nndtw}, 1NN-DTW is able to detect some of the error patterns with accuracies of more than 60 percent. This is comparable to the results by \citeauthor{giggins2014rehabilitation}~\cite{giggins2014rehabilitation} and \citeauthor{o2015evaluating}~\cite{o2015evaluating} for simple rehabilitation exercises. The computational cost of DTW are quadratic with respect to the lengths of the trajectories. In our setting, a single DTW takes about \SI{55}{ms} on average per trajectory. On average, the trajectories used for this experiment consist of 500 frames. For each trajectory to be classified, DTW has to be calculated with each of our training trajectories ($N=95$). This leads to an average time of over 5 seconds to calculate the DTWs necessary for one single query trajectory. Thus, even if the classification led to optimal results, it would not be applicable in an interactive setting.

\subsection{Reducing Alignment Cost: 1NN-RefDTW}
\label{sec:1nnrefdtw}
To reduce computational cost, we can exploit the general similarity between the trajectories that all represent the same motor action (squat). We can thus warp all training trajectories  to a normalized timing in an offline preprocessing step. This is done by selecting one reference trajectory $T_r$ and warping all trajectories to its timing. If $T_r$ it is a very short trajectory (i.e.\ a fast movement), information from the original trajectories gets lost due to the warping. Thus, as reference trajectory, we select the longest trajectory that contains all available movement segments. The warping exploits the correspondences found by DTW. For each frame $t$ of $T_r$, the corresponding frame in the to-be-warped trajectory is selected according to the correspondence path from DTW. 

For classification, we perform 1NN using the mean of the frame-by-frame distance between the warped query trajectory $T_q$ and the warped training trajectories as distance measure:
\begin{equation*}
 dist(T_q,T_i) = \frac{1}{\lvert T_r \rvert}\sum_{t=1}^{\lvert T_r \rvert} \sum_{d=1}^{k} (1- |\vec{q}_{t,d}^q \vec{q}_{t,d}^i|).
\end{equation*}
$\vec{q}_{t,d}^q$ is the quaternion describing the $d$-th joint in the $t$-th frame of the warped query trajectory, whereas $\vec{q}_{t,d}^i$ refers to the corresponding joint of the training trajectory $i$. $\lvert T_r \rvert$ is the length of the reference trajectory. In our case, we have $\lvert T_r \rvert=902$. For each classification, the calculation of one DTW for $T_q$ is sufficient: All comparisons between warped query and training trajectories can now be done frame-by-frame with computational cost linear in $\lvert T_r \rvert$. In our setting, this process needs on average \SI{25}{ms} per trajectory. We call the resulting algorithm \textit{1NN-RefDTW} and expect it to have similar classification performance as 1NN-DTW while incurring reduced computational cost.

\begin{figure}[] 
 \centering
 \centerline{ 
 		\includegraphics[width=1.0\columnwidth]{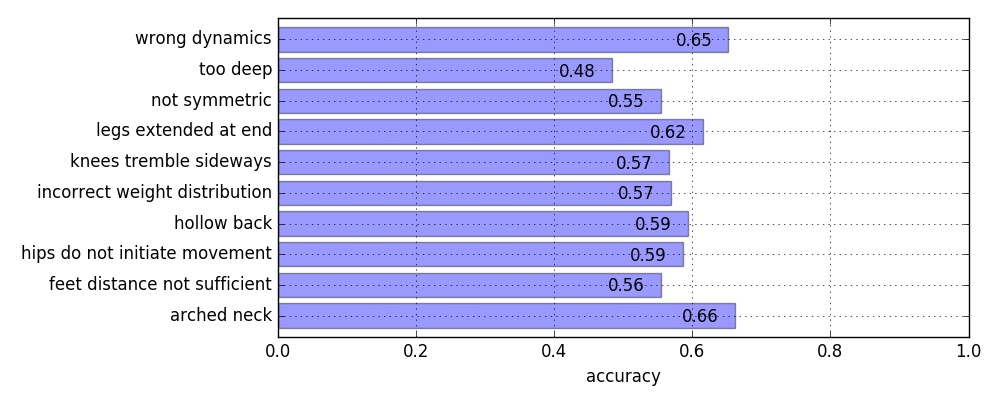}
 } 
 \caption{Accuracy results for classifier 1NN-RefDTW.}
 \label{fig:res:1nnRefDTW}
 \end{figure}

Figure~\ref{fig:res:1nnRefDTW} summarizes the classification results of 1NN-RefDTW. The classification accuracy is comparable to 1NN-DTW, with some error patterns detected slightly better. Still, the classification accuracy is insufficient for being applied in a coaching scenario. Concerning the computational costs, the new classifier only needs one DTW per query trajectory. Warping a training trajectory into the timing of the reference trajectory needs on average \SI{90}{ms}. Additionally, the frame-to-frame distance between the warped query and the training trajectories has to be calculated. The computational effort for classification is thus $|T|^2+N|T|$ instead of $N|T|^2$ if all trajectories are of size $|T|$. In our setting, the classification process for $N=95$ needs approximately \SI{2.5}{s}. However, the time needed for classification still depends on the number of trajectories in the data set, which is problematic for large training sets. 

\subsection{Separate Classification of Error Patterns: RefDTW-SVM}
\label{sec:dtwrefsvm}
Errors during the performance of motor actions can occur in many different combinations. 1NN-DTW and its extension 1NN-RefDTW only return the whole set of labels of the nearest neighbor as classification for each query. Combinations of error patterns that do not exist in the training data cannot be detected by the algorithm, unless the training data contains all possible combinations of error patterns. As this is typically not the case, it is desirable to learn a separate classifier for each pattern. Furthermore, we would like to provide a classifier with even more reduced computational cost, ideally independent of the size of the training set.
Both goals can be achieved using Support Vector Machines (SVM), one of the most successful machine learning algorithms in general~\cite{fernandez2014we}. An SVM learns a decision hyperplane which maximizes the margin between two classes~\cite{bishop2006pattern}. For classification, the SVM only has to determine on which side of a hyperplane an input query lies. In our case, we can learn a classifier for each error pattern, considering each training trajectory as one data point with the label \textit{pattern occurs} or \textit{pattern does not occur}. To use the SVM for training, we first warp all training trajectories to the timing of the reference trajectory. Then, for each warped training trajectory, a feature vector is constructed and standardized via scaling to unit variance and removing the mean. This vector consists of all joint angles in Euler angle representation as well as the joint positions for each frame in the warped trajectory. The feature vector thus has size $6  \lvert T_r \rvert  k$, where $\lvert T_r \rvert$ is the number of frames of the reference trajectory and $k$ the number of joints. In our case, we have $\lvert T_r \rvert=902$ and $k=19$. Again, we tested different features and found that using joint angles in Euler angle representation together with joint positions leads to good classification results (cf. supplementary online material).

We trained one two-class SVM for each error pattern on the feature vectors obtained from the warped trajectories. In our experiments, a non-linear RBF kernel was unable to beat the linear kernel, thus we decided to use SVMs with linear kernel (cf. supplementary online material). We use the standard SVM implementation from scikit-learn~\cite{scikit-learn} in version 0.17.1. For classification, a query trajectory is first warped to the timing of the reference trajectory. Then the feature vector is constructed and classified by the trained SVMs. The resulting algorithm is called RefDTW-SVM.

\begin{figure}[] 
 \centering
 \centerline{ 
 	\includegraphics[width=1.0\columnwidth]{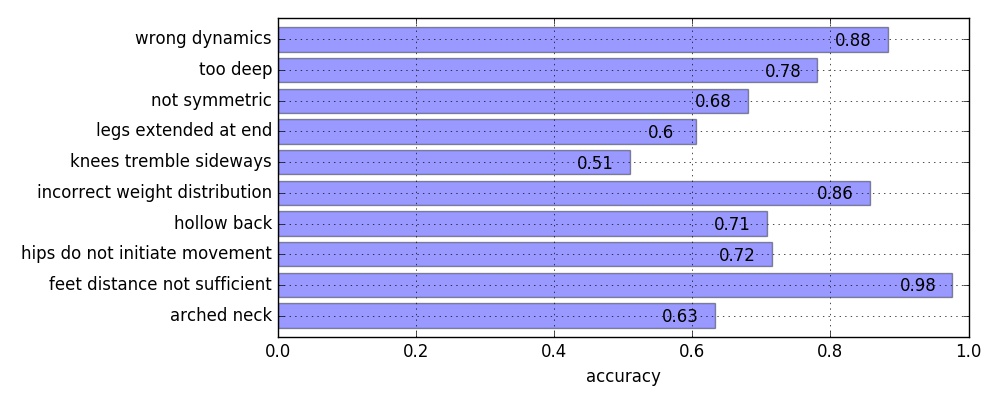}
 } 
 \caption{Accuracy results for classifier RefDTW-SVM using joint angles as well as joint positions as features.}
 \label{fig:res:1nnRefDTWsvmallfeatures}
 \end{figure}

Results can be seen in Figure~\ref{fig:res:1nnRefDTWsvmallfeatures}: Now, three of the error patterns are classified with an accuracy greater than \SI{80}{\%}. Also, most of the other patterns reach higher results than with the previous 1NN approaches. However, the overall classification performance is still not sufficient. One explanation is the immense number of features per trajectory. We will approach this problem in the next section. Concerning the time needed for classification, for each error pattern, the classifier now only needs a mean of \SI{9.7}{ms}. Before starting the classification of error patterns, one DTW has to be calculated, which takes about \SI{90}{ms} as described in Section \ref{sec:1nnrefdtw}.

\subsection{Reducing Features: RefDTW-RF-SVM}
Our feature vector of size $6 \lvert T_r \rvert k$ comprises many irrelevant features: For instance, we intuitively do not consider the rotation of the wrist to be related to having a straight back. The SVM classifier might suffer from this high number of irrelevant features as shown by \citeauthor{weston2000feature}~\cite{weston2000feature} and \citeauthor{chen2006combining}~\cite{chen2006combining}. According to their results, we assume a robust feature selection method to be able to help improving classifier performance. To this end, we use Random Forests (RF) for feature selection~\cite{genuer2010variable,chen2006combining}. Random Forests perform feature selection as well as classification. They are based on Decision Trees, which learn a hierarchical set of rules to distinguish between classes. Thereby, they implicitly weight the importance of each feature. Random Forests extend Decision Trees and reduce their susceptibility to overfitting via training multiple randomized Decision Trees and averaging them. This leads to an improved accuracy of the estimator as well as a reduced overfitting~\cite{breiman2001random}. See~\cite{biau2012analysis} for an in-depth analysis of the statistical properties and the mathematical background of Random Forests.

Direct classification using Random Forests leads to high computational cost, as all trees in the forest must be considered. We are interested in a model that provides good classification performance with minimal time for classification. As the SVM-based classification presented in Section \ref{sec:dtwrefsvm} provides almost acceptable results in real-time, we boosted it with a Random-Forest-based feature selection: We trained one Random Forest for each error pattern. The Random Forests are trained on the same feature vectors extracted from the warped trajectories as described for RefDTW-SVM. To train the trees, we used the Gini impurity as criterion to optimize the decision rules. As break condition for growing, we require all leaves to contain only a single class or less than two samples. We observed a number of 200 trees to lead to good results. 

The idea of our new algorithm RefDTW-RF-SVM is to use the Random Forests only for feature selection during training: For each error pattern, the Random Forest assigns an importance value to each feature via averaging the relative importance of the feature in each decision tree. Following an idea of \citeauthor{bi2003dimensionality}~\cite{bi2003dimensionality}, we add 20 random features to each frame before performing the feature weighting by Random Forests. The average of their importance values is used as threshold to discard irrelevant features. This leads to 580 features on average per error pattern (from originally around 100,000 features) which we use as input for the SVMs. We trained the SVMs with the same parameters as for RefDTW-SVM. For the Decision Trees as well as the Random Forests, we use the the scikit-learn implementation~\cite{scikit-learn}.

\begin{figure}[] 
 \centering
 \centerline{ 
 		\includegraphics[width=1.0\columnwidth]{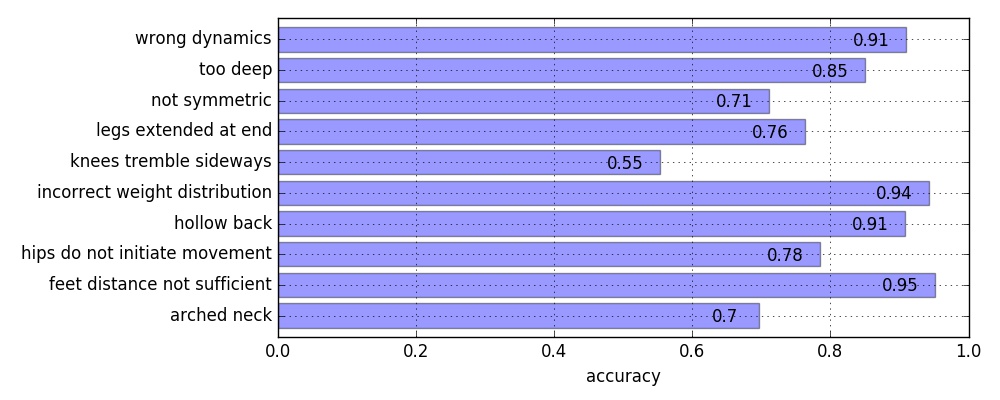}
 } 
 \caption{Accuracy results for classifier RefDTW-RF-SVM.}
 \label{fig:res:RefDTWrfsvm}
 \end{figure}

Figure~\ref{fig:res:RefDTWrfsvm} shows the resulting classification accuracy, which outperforms RefDTW-SVM for nearly all patterns. Five patterns reach accuracies higher than 80 percent. Concerning the classification time, only \SI{0.1}{ms} is needed in addition to the DTW step. This leads to a total time to classify all patterns after DTW of around \SI{1}{ms}. 

\subsection{Getting Classification Results Earlier: Segment-based RefDTW-RF-SVM}
\label{sec:finalAlgorithm}
RefDTW-RF-SVM and all other approaches presented before only allow classification after the whole motor action is completed, as the full query trajectory needs to be warped by DTW. However, some error patterns are limited to parts of the motor action. For instance, the desired depth of the squat is relevant only at the deepest point of the motion. This information can be exploited by using the concept of \textit{movement segments}: Each performance of a motor action can be considered a combination of simpler sequential sub-actions. These movement segments are homogeneous and functionally meaningful parts of a more complex movement. For the squat, we define the movement segments \textit{preparation}, \textit{going down}, \textit{is down}, \textit{going up}, and \textit{wrap up}. 

The underlying idea of a segment-based RefDTW-RF-SVM is to simply apply RefDTW-RF-SVM to a single movement segment once it has been completed. The segmentation is done based on a state machine which splits the trajectory at boundary points (state changes) where important joints like the knees start or stop moving. This is similar to the approach proposed in~\cite{hulsmann2016multi}. The segmentation takes less than \SI{1}{ms} per frame.

\begin{figure}[] 
 \centering
 \centerline{ 
 		\includegraphics[width=1.0\columnwidth]{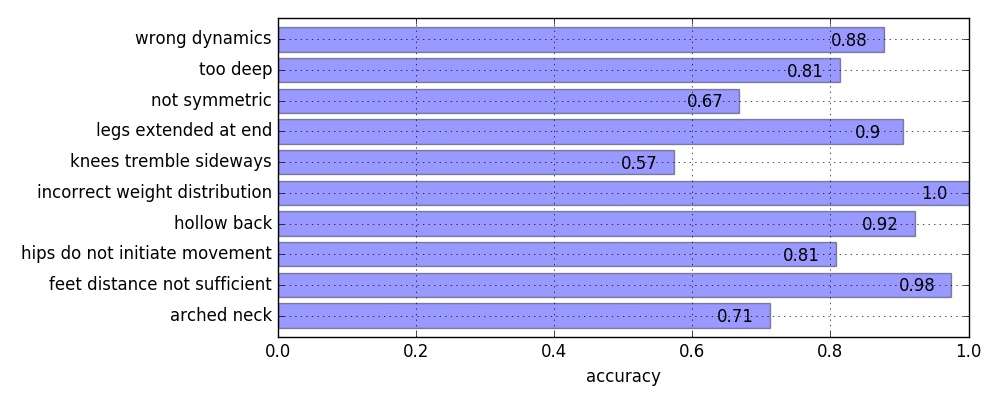}
 } 
 \caption{Accuracy of classifier Segment-based RefDTW-RF-SVM.}
 \label{fig:res:RefDTWrfsvmms}
 \end{figure}

As shown in Figure~\ref{fig:res:RefDTWrfsvmms}, the classification results are comparable to the results obtained with RefDTW-RF-SVM, which however works on the complete trajectories. For each pattern, the maximum accuracy per movement segment is reported. Seven error patterns are classified with an accuracy of above 80 percent. We performed the classification with the automatically segmented trajectories as well as with manually segmented trajectories. Both led to similar results. Concerning the time needed for classification, as the trajectories for the movement segments are shorter than for the whole motor actions, DTW only needs about \SI{10}{ms} per movement segment instead of about \SI{90}{ms} for a whole trajectory. The classification step itself using Segment-based RefDTW-RF-SVM needs around \SI{0.1}{ms}. Overall, an error pattern is classified on average around \SI{10.1}{ms} after the movement segment of interest has been performed. As the DTW, which is responsible for around \SI{10}{ms} of this time, has to be performed only once, we now need approximately \SI{11.0}{ms} to classify each of our ten error patterns.

\subsection{Summary of the Results}
\label{ref:results}
\begin{figure}[] 
 \centering
 \centerline{ 
 		\includegraphics[width=1.0\columnwidth]{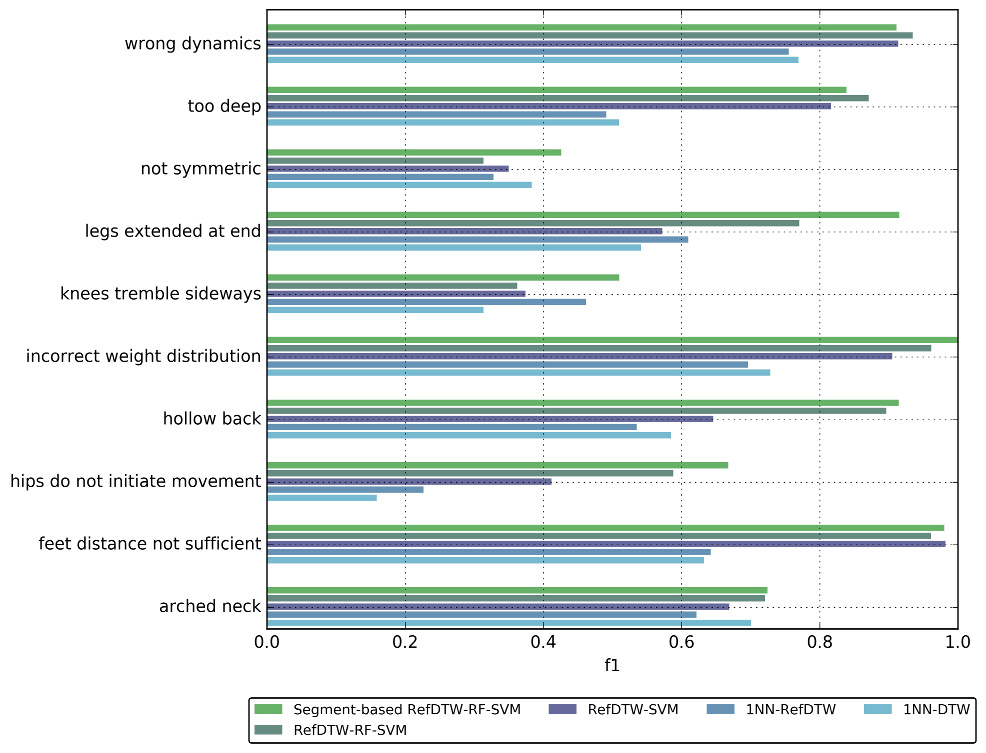}
 } 
 \caption{F1 Score of all classifiers.}
 \label{fig:classificationF1}
 \end{figure}

All algorithms except from 1NN-RefDTW were able to beat the classification performance of our baseline 1NN-DTW. The best classification quality is achieved by RefDTW-RF-SVM and Segment-based RefDTW-RF-SVM.  
Most error patterns, including the most frequent ones ``wrong dynamics'', ``incorrect weight distribution'', and ``too deep'', can all be detected with an accuracy above \SI{80}{\%}. The patterns ``incorrect weight distribution'' and ``feet distance not sufficient'' are even nearly  perfectly classified. Only the error patterns with the fewest occurances in our training data, namely ``not symmetric'' (17 occurances) and ``knees tremble sideways'' (23 occurances) are classified with an accuracy below \SI{70}{\%}. Additionally, Figure~\ref{fig:classificationF1} reports the F1 score of all presented approaches. Concerning the F1 score, the data looks similar: Only four patterns are classified with a score below 0.8. This enables our system to make use of various feedback strategies (cf. video in the supplementary online material). The exact scores and their standard deviation in the 5-fold cross validation can be found in the supplementary online material.
All algorithms, except from Segment-based RefDTW-RF-SVM, require the calculation of DTW on the whole trajectory, which takes on average about \SI{90}{ms}. Segment-based RefDTW-RF-SVM only needs single movement segments to be warped, which can be performed in around \SI{10}{ms}. Table~\ref{tab:time} summarizes the time needed to classify a query trajectory with respect to the ten error patterns. Segment-based RefDTW-RF-SVM is clearly the fastest classifier as the classification step itself only needs \SI{1}{ms} and the result is potentially available already during the execution of the movement, directly after a single movement segment has been completed.
 
 \begin{table}
 \center
  \caption{Time needed to classify a query trajectory with respect to all ten error patterns, including the time needed for DTW. The reference-based DTW takes \SI{89}{ms} for a complete trajectory and on average \SI{10}{ms} on movement segments. For the 1NN-based approaches, all 95 training trajectories were used for classification.}%
  \label{tab:time}
  \begin{minipage}{\columnwidth}
\begin{center}

\begin{tabular}{m{2.2cm}m{1.5cm}m{1.5cm}m{1.5cm}m{1.5cm}m{2.5cm}}
  \hline
 &1NN-DTW & 1NN-RefDTW  &RefDTW-SVM  & RefDTW-RF-SVM & Segment-based RefDTW-RF-SVM \\
 \hline
    \begin{tabular}{c} Time in ms \end{tabular} & $>$~\SI{5000}{ms} & approx. \SI{2500}{ms} &  \SI{187}{ms}  &  \SI{90}{ms} & \SI{11}{ms}  \\ 
\hline
\end{tabular}
\end{center}
\end{minipage}
\end{table}

\section{Discussion and Conclusion}
\label{sec:discussion}
We have presented steps to yield a novel classifier for a fast detection of a variety of error patterns in movement trajectories, as required for interactive coaching applications, e.g., in virtual reality environments. We evaluated all algorithms on a complex motor task involving a high number of relevant error patterns. All scores were measured using cross validation, in a setup where data from one single subject is not allowed to be distributed over multiple folds. Thus our results capture the algorithms' abilities to generalize across subjects. The resulting algorithm, Segment-based RefDTW-RF-SVM, provides the best balance between quality of classification and computation time: Besides being the fastest classifier in our set, it is among the two classifiers with the highest accuracy scores. Nearly all error patterns, especially the most frequent ones, are classified with accuracies above \SI{80}{\%}. In contrast to many related approaches, this classifier is able to work in interactive setups as shown in our demonstration of how online verbal feedback can be triggered through our automatic error analysis (see the video in the supplementary material). 

Overall, from the evaluation of each of the different steps taken in the previous section, we can derive the following conclusions about automatic error analysis of human motor performances:
\begin{enumerate}
   \item If the data consists of structurally similar movements such as the same type of motor actions, it is sufficient to temporally align all trajectories via performing DTW with one reference trajectory. Thereby we were able to reduce the computational effort while keeping the quality of the classification in a similar range for nearly all error patterns.
   \item For the classification of multiple error patterns, independent classifiers should be trained. A nearest neighbor-based classification, which only copies all labels from the nearest neighbor of a query, is insufficient especially for small training data sets. Learning independent classifiers for all error patterns increased the classifier performance for nearly all examined error patterns.
   \item Random Forests help to select relevant features from high-dimensional input trajectories, even if the number of training examples is small. Such a preprocessing step significantly improves the performance of SVM-based classification. This holds especially for error patterns which are characterized only by very few features such as the ``hollow back''.
   \item By classifying data from appropriate movement segments, instead of whole trajectories, the time needed for classification can be drastically minimized while keeping the classification performance high.
\end{enumerate}

Note that even though general classification performance of our algorithm is high, the performance is not convincing specifically for two error patterns: The pattern ``not symmetric'' is detected only with F1 scores around 0.43. This error pattern is annotated in trajectories where some joints are not symmetric between the left and the right side of the body. As this can occur in almost all joints and all phases of the movement, the feature selection cannot easily spot those features of interest that are relevant. Further, the classifier has no possibility to infer information on the relationship between multiple joints with respect to symmetry. For the other problematic pattern, ``knees tremble sideways'', our best classifier only achieves an F1 score of 0.51. This pattern describes a very subtle movement. Also, it can spread temporarily: Exactly the frames that are problematic for subject A can be correct for subject B and vice versa. Finally, the number of trembles can be different for different subjects which also makes classification harder. One way to deal with these two  problematic patterns is the construction of more complex higher-level features. A higher-level feature could, for instance, describe the relationship between certain parts of the body or the movement of the athlete's center of mass. The automatic generation and inclusion of such higher-level features is a promising field of future work. Another limitation is that  temporal properties of the movements are not covered directly by our algorithm. For motor actions where the user's timing has an influence on whether certain errors occur, temporal information could be included via adding velocity as well as information on the warping function extracted from DTW. 

\section{Acknowledgements}
This research was supported by the Cluster of Excellence Cognitive Interaction Technology 
CITEC (EXC 277) at Bielefeld University, which is funded by the German Research Foundation (DFG).

\bibliography{analysis-of-motor-performance}

\appendix

\newpage
\section{Supplementary Online Materials}
\subsection{Detailed Scores}
\begin{figure}[h] 
 \centering
 \centerline{ 
 		\includegraphics[width=0.8\columnwidth]{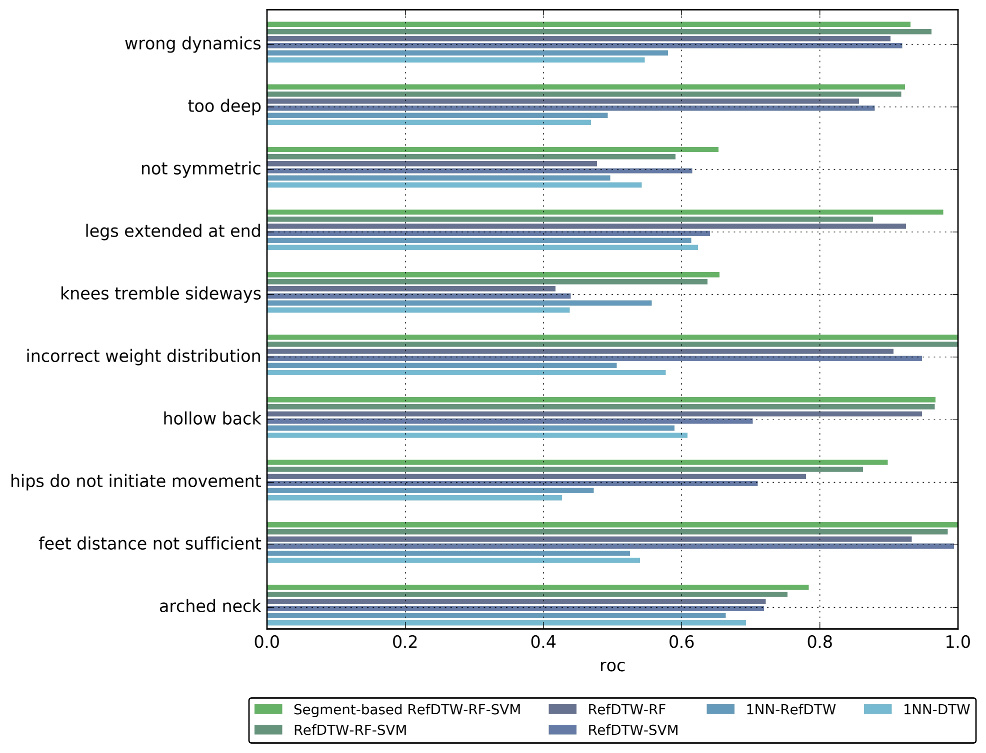}
 } 
 \caption{ROC AUC Score of all classifiers.}
 \label{fig:classificationRoc}
 \end{figure}

\begin{figure}[h] 
 \centering
 \centerline{ 
 		\includegraphics[width=0.8\columnwidth]{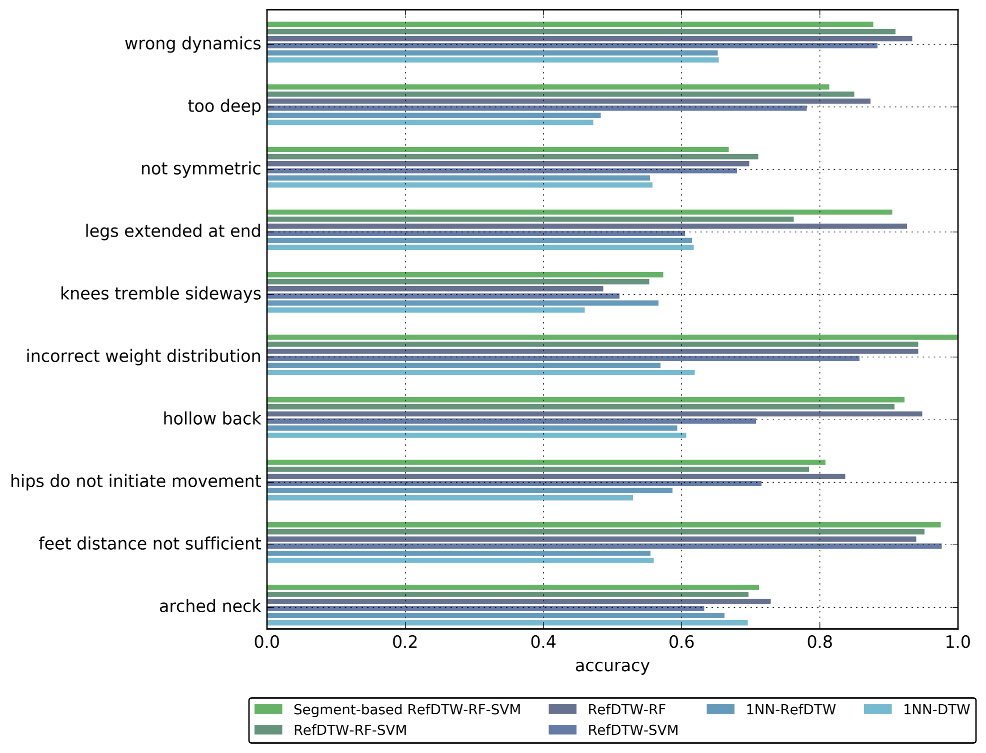}
 } 
 \caption{Accuracy of all classifiers.}
 \label{fig:classificationAcc}
 \end{figure}

Here, we report the measured classification performance of all tested classifiers with respect to accuracy (see Figure~\ref{fig:classificationAcc} and Table~\ref{tab:classificationAcc}), F1 score (see Table~\ref{tab:classificationF1}) and Receiver Operating Characteristics Area Under the Curve (ROC AUC) (see Figure~\ref{fig:classificationRoc} and Table~\ref{tab:classificationRoc}). ROC curves provide a plot which describes the relationship between recall and fall-out. The true positive rate is plotted on the y axis, the false positive rate on the x axis. The higher the curve, the better the classification. The area under the curve (AUC) is thus often used as score for classifier performance as it provides the probability to rank a randomly chosen positive instance higher than a randomly chosen negative one. Thus the higher the result, the better the classifier performs.
This section also contains results for the pure Random-Forest-based classification (RefDTW-RF). This, leads to a classification performance in a similar range to RefDTW-RF-SVM, but also to more computational effort: We need around \SI{160}{ms} additional to the time needed for the DTW step to classify all of our 10 error patterns even if the trees inside the Random Forests are evaluated in parallel. All further components of the system, such as dialogue planning, Text-to-Speech, coaching animation, et cetera have to wait this period of time until they can start planning the feedback corresponding to the motion the trainee just performed in the virtual environment. Thus, for RefDTW-RF-SVM, we only use the Random Forests for feature selection during training to significantly speed up the classification time.

\subsection{Comparison of Different Feature Types}
\begin{figure}[] 
 \centering
 \centerline{ 
 		\includegraphics[width=0.8\columnwidth]{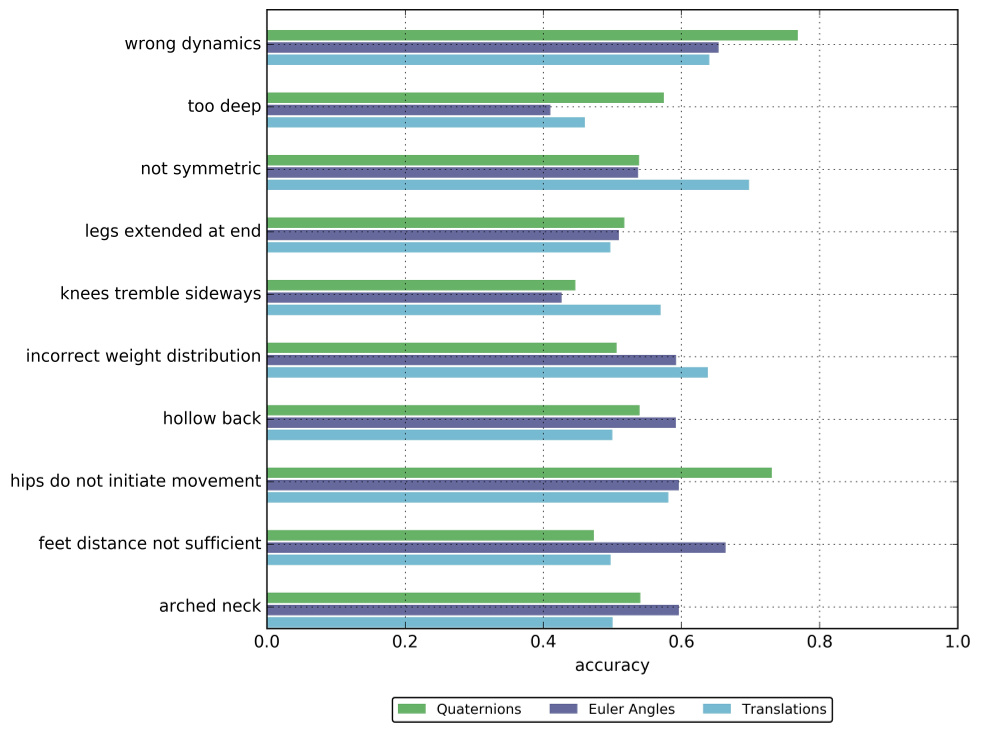}
 } 
 \caption{Accuracies for different types of features for classifier 1NN-DTW.}
  \label{fig:1nndtw:features:acc}
 \end{figure}

\begin{figure}[] 
 \centering
 \centerline{ 
 		\includegraphics[width=0.8\columnwidth]{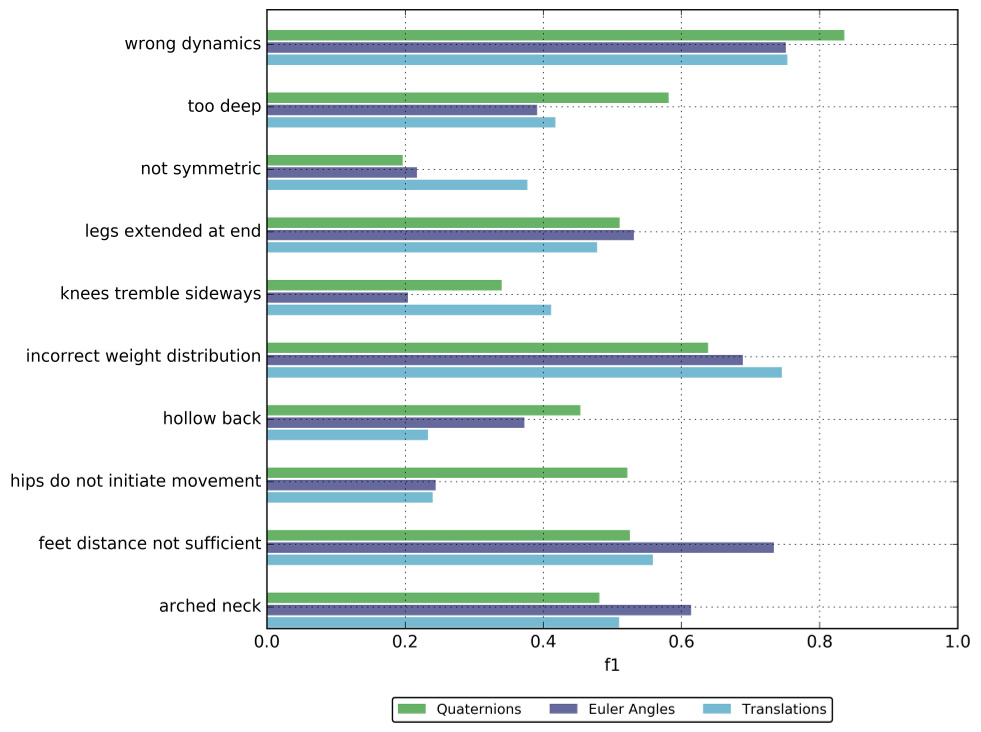}
 } 
 \caption{F1 scores for different types of features for classifier 1NN-DTW.}
  \label{fig:1nndtw:features:f1}
 \end{figure}

First, we compare the classification results of the baseline 1NN-DTW when using rotations as quaternions, as Euler angles or using joint positions. In the nearest neighbor step, the euclidean distance between the warped frames is used. All approaches lead to results in a similar range on average over all error patterns (see Figure~\ref{fig:1nndtw:features:acc} and Figure~\ref{fig:1nndtw:features:f1}).

\begin{figure}[] 
 \centering
 \centerline{ 
 		\includegraphics[width=0.8\columnwidth]{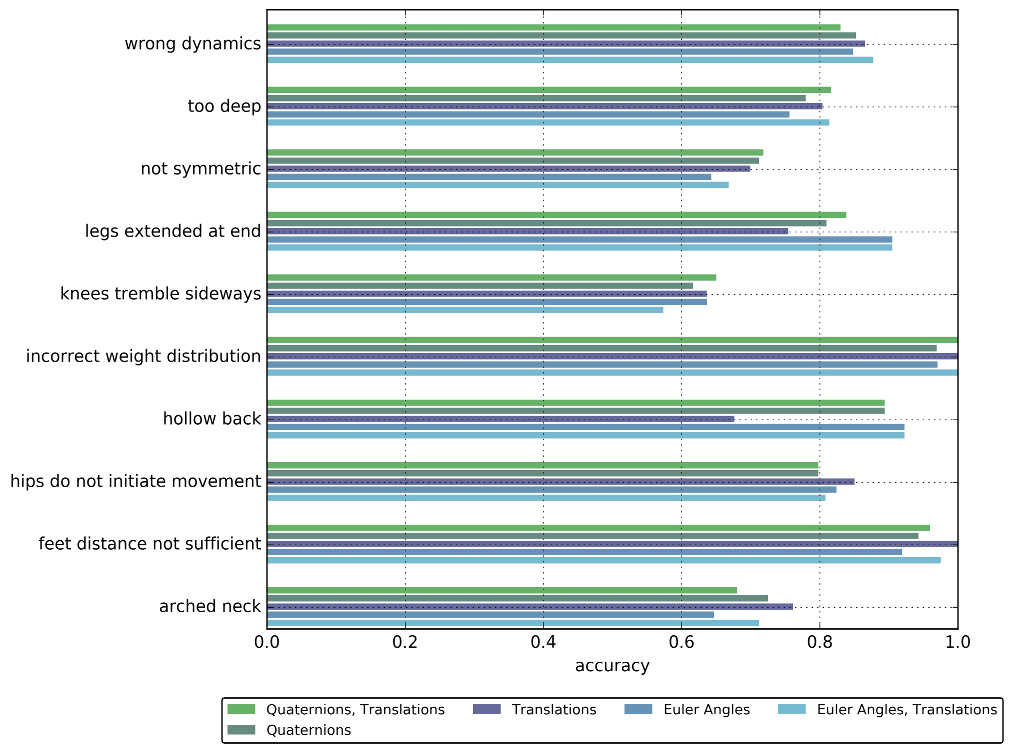}
 } 
 \caption{Accuracies for different types of features for classifier Segment-based RefDTW-RF-SVM.}
  \label{fig:final:features:acc}
 \end{figure}

\begin{figure}[] 
 \centering
 \centerline{ 
 		\includegraphics[width=0.8\columnwidth]{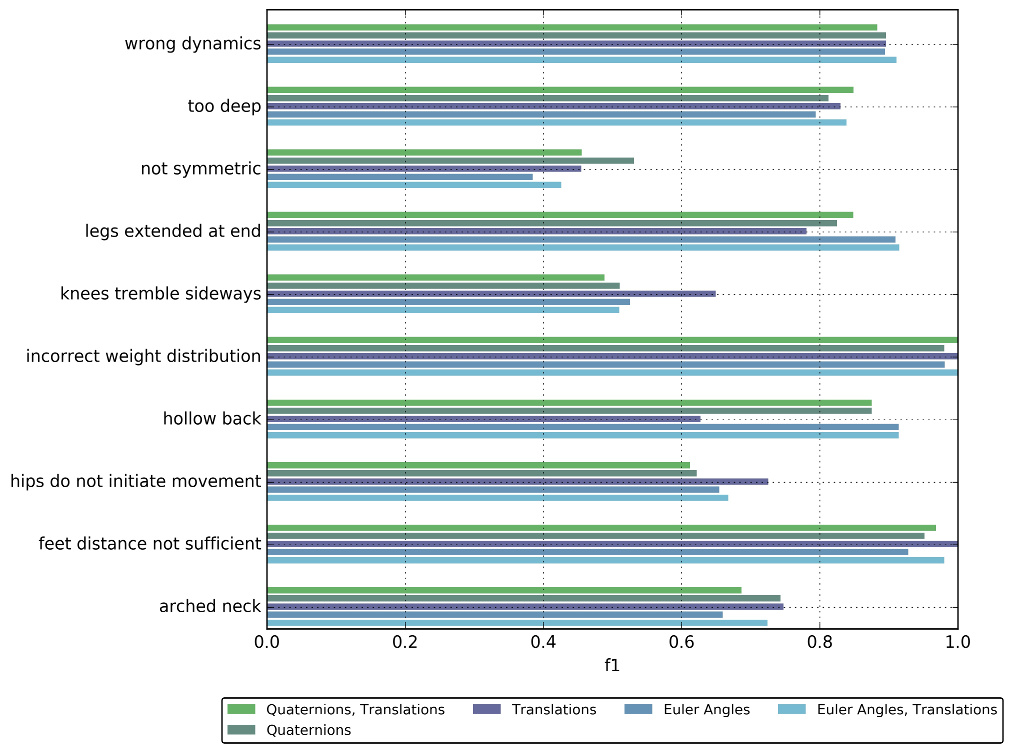}
 } 
 \caption{F1 scores for different types of features for classifier Segment-based RefDTW-RF-SVM.}
  \label{fig:final:features:f1}
 \end{figure}

Second, we compare the classification results of our own final classifier Segment-based RefDTW-RF-SVM with respect to different feature sets. The feature weighting using Random Forests on quaternions is implemented component-wise. All quaternions with at least one feature weight above the threshold are completely used for SVM classification. For some error patterns, we observe that the quality of the classification complements each other for joint angles and joint translations: Some patterns (such as ``feet distance not sufficient'') can be classified best based on the translations, others (such as ``hollow back'') are classified much better based on the angles. We thus combine joint angles and joint translations which leads to a slight enhancement of the overall performance. Here, we finally decide to use Euler angles instead of quaternions for the sake of better interpretability of the selected features and a slightly shorter feature vector. In general, all classifiers behave similarly (see Figure \ref{fig:final:features:acc} for the accuracies and Figure \ref{fig:final:features:f1} for the F1 scores). 

\subsection{RFB Kernel vs. Linear Kernel in SVM}
\begin{figure}[] 
 \centering
 \centerline{ 
 		\includegraphics[width=0.8\columnwidth]{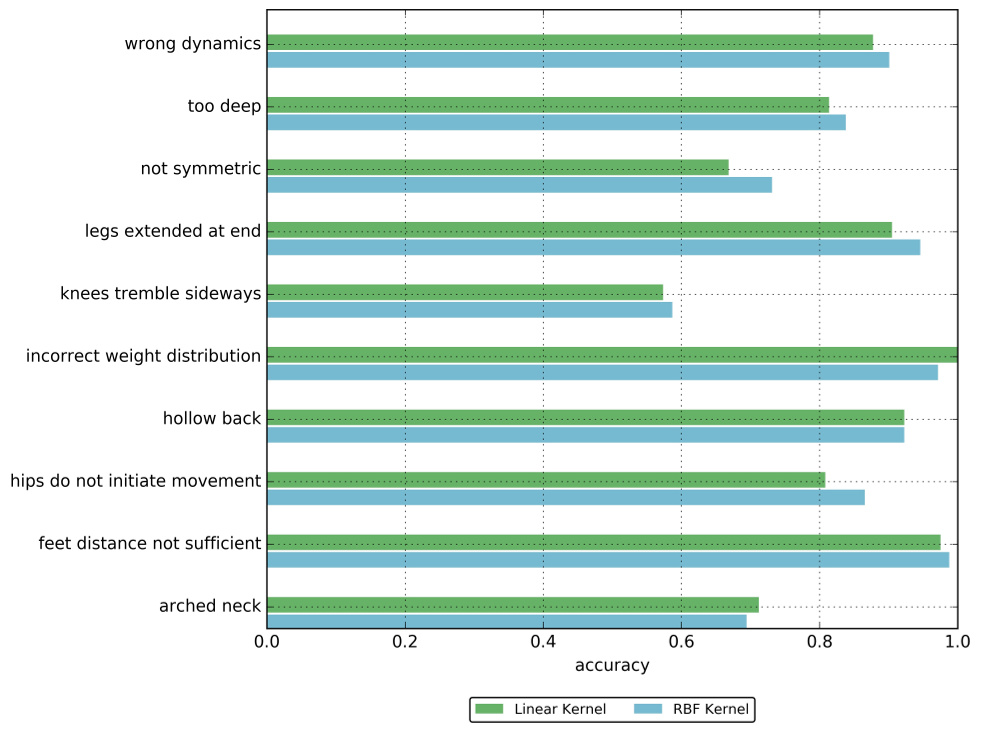}
 } 
 \caption{Accuracies for linear and rbf kernel for classifier Segment-based RefDTW-RF-SVM.}
  \label{fig:final:rbfvslinear:acc}
 \end{figure}

\begin{figure}[] 
 \centering
 \centerline{ 
 		\includegraphics[width=0.8\columnwidth]{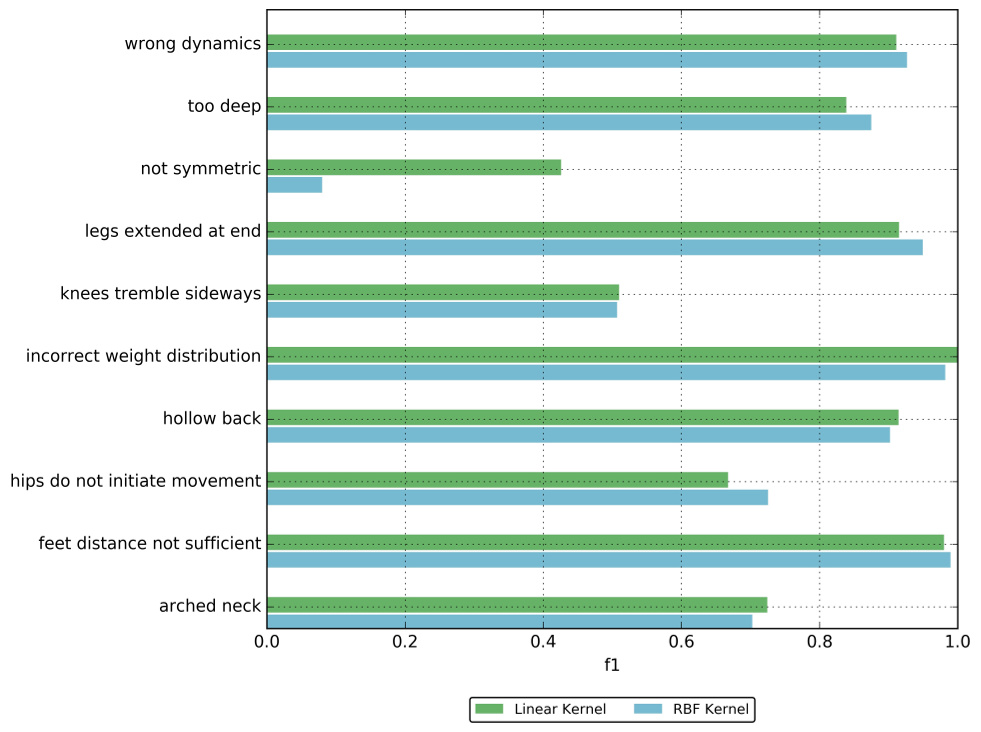}
 } 
 \caption{F1 scores for linear and rbf kernel for classifier Segment-based RefDTW-RF-SVM.}
  \label{fig:final:rbfvslinear:f1}
 \end{figure}

In this part, we compare the classification performance of Segment-based RefDTW-RF-SVM using a linear kernel compared to using a radial basis function kernel. Results are in a similar range (see Figure \ref{fig:final:rbfvslinear:acc} for the accuracies and Figure \ref{fig:final:rbfvslinear:f1} for the F1 scores). We finally decide to use the linear kernel for the sake of simplicity.

\subsection{1NN-DTW Based on Movement Segments}
\begin{figure}[] 
 \centering
 \centerline{ 
 		\includegraphics[width=0.8\columnwidth]{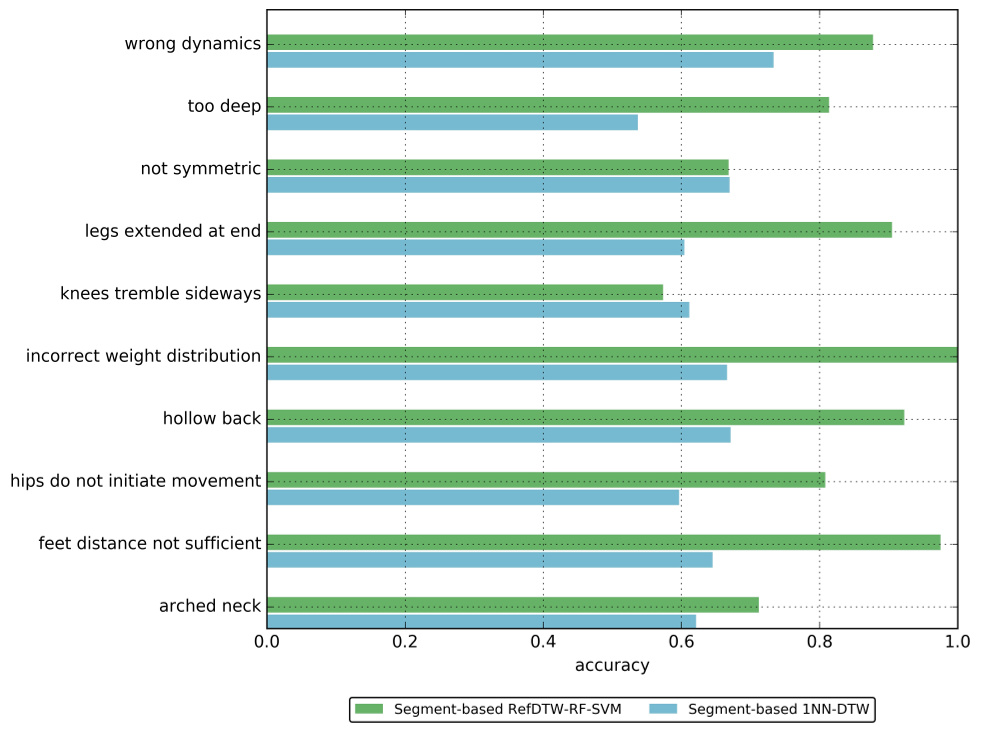}
 } 
 \caption{Accuracies for the comparison of baseline 1NN-DTW on movement segments with Segment-based RefDTW-RF-SVM.}
  \label{fig:final:1nndtwsegment:acc}
 \end{figure}

\begin{figure}[] 
 \centering
 \centerline{ 
 		\includegraphics[width=0.8\columnwidth]{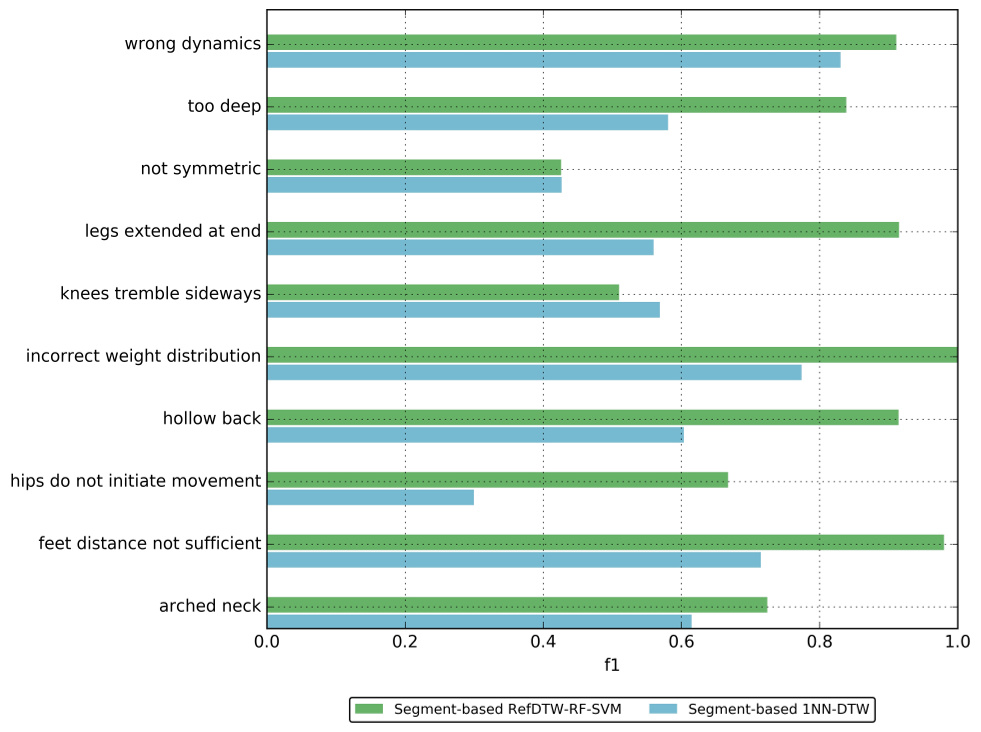}
 } 
 \caption{F1 scores for the comparison of baseline 1NN-DTW on movement segments with Segment-based RefDTW-RF-SVM.}
  \label{fig:final:1nndtwsegment:f1}
 \end{figure}

Finally, we evaluated the performance of 1NN-DTW on movement segments which leads to Segment-based 1NN-DTW. Here, the results are again worse than for our own classifier Segment-based RefDTW-RF-SVM (see Figure \ref{fig:final:1nndtwsegment:acc} for the accuracies and Figure \ref{fig:final:1nndtwsegment:f1} for the F1 scores).

 \newcommand{\score}{accuracy}
\setcounter{pidx}{1}
\begin{table}%
\caption{Classification performance: Accuracy.}
\label{tab:classificationAcc}
\begin{minipage}{\columnwidth}
\begin{center}
\begin{tabular}{m{3.6cm}m{1.4cm}m{1.4cm}m{1.4cm}m{1.4cm}m{1.4cm}m{1.4cm}}
\hline
\restableheader{Accuracy \\ \textit{(SD)}}
\resline{\score}
\resline{\score}
\resline{\score}
\resline{\score}
\resline{\score}
\resline{\score}
\resline{\score}
\resline{\score}
\resline{\score}
\resline{\score}
\end{tabular}
\end{center}
\end{minipage}
\end{table}%

\renewcommand{\score}{f1}
\setcounter{pidx}{1}
\begin{table}%
\caption{Classification performance: F1 Score.}
\label{tab:classificationF1}
\begin{minipage}{\columnwidth}
\begin{center}
\begin{tabular}{m{3.6cm}m{1.4cm}m{1.4cm}m{1.4cm}m{1.4cm}m{1.4cm}m{1.4cm}}
\hline
\restableheader{F1 \\ \textit{(SD)}}
\resline{\score}
\resline{\score}
\resline{\score}
\resline{\score}
\resline{\score}
\resline{\score}
\resline{\score}
\resline{\score}
\resline{\score}
\resline{\score}
\end{tabular}
\end{center}
\end{minipage}
\end{table}%

\renewcommand{\score}{roc}
\setcounter{pidx}{1}
\begin{table}%
\caption{Classification performance: ROC AUC Score.}
\label{tab:classificationRoc}
\begin{minipage}{\columnwidth}
\begin{center}
\begin{tabular}{m{3.6cm}m{1.4cm}m{1.4cm}m{1.4cm}m{1.4cm}m{1.4cm}m{1.4cm}}
\hline
\restableheader{ROC AUC \\ \textit{(SD)}}
\resline{\score}
\resline{\score}
\resline{\score}
\resline{\score}
\resline{\score}
\resline{\score}
\resline{\score}
\resline{\score}
\resline{\score}
\resline{\score}
\end{tabular}
\end{center}
\end{minipage}
\end{table}%

\end{document}